\pdfoutput=1

\documentclass[11pt]{article}

\usepackage{emnlp2021}
\usepackage{times}
\usepackage{latexsym}
\usepackage{amsmath}
\usepackage{multicol}
\DeclareMathOperator*{\argmax}{arg\,max}
\usepackage{pifont}

\usepackage[T1]{fontenc}

\usepackage[utf8]{inputenc}

\usepackage{microtype}
\newcommand{\cmark}{\ding{51}}%
\newcommand{\xmark}{\ding{55}}%

\title{NVIDIA NeMo's Neural Machine Translation Systems for English $\leftrightarrow$ German and English $\leftrightarrow$ Russian News and Biomedical Tasks at WMT21}

\author{Sandeep Subramanian, Oleksii Hrinchuk, Virginia Adams, Oleksii Kuchaiev \\
         NVIDIA \\ Santa Clara, CA \\ 
         \textit{\{sandeepsub, ohrinchuk, vadams, okuchaiev\}@nvidia.com}}

\begin{document}
\maketitle
\begin{abstract}
This paper provides an overview of NVIDIA NeMo's neural machine translation systems for the constrained data track of the WMT21 News and Biomedical Shared Translation Tasks. Our news task submissions for English $\leftrightarrow$ German (En $\leftrightarrow$ De)  and English $\leftrightarrow$ Russian (En $\leftrightarrow$ Ru) are built on top of a baseline transformer-based sequence-to-sequence model \cite{vaswani2017}. Specifically, we use a combination of 1) checkpoint averaging 2) model scaling 3) data augmentation with backtranslation and knowledge distillation from right-to-left factorized models 4) finetuning on test sets from previous years 5) model ensembling 6) shallow fusion decoding with transformer language models and 7) noisy channel re-ranking. Additionally, our biomedical task submission for English $\leftrightarrow$ Russian uses a biomedically biased vocabulary and is trained from scratch on news task data, medically relevant text curated from the news task dataset, and biomedical data provided by the shared task. Our news system achieves a sacreBLEU score of 39.5 on the WMT'20 En $\rightarrow$ De test set outperforming the best submission from last year's task of 38.8. Our biomedical task Ru $\rightarrow$ En and En $\rightarrow$ Ru systems reach BLEU scores of 43.8 and 40.3 respectively on the WMT'20 Biomedical Task Test set, outperforming the previous year's best submissions. 
\end{abstract}

\section{Introduction}
We take part in the WMT'21 News Shared Task for English $\leftrightarrow$ German, English $\leftrightarrow$ Russian, and the Biomedical Shared Task for English $\leftrightarrow$ Russian. Our systems are implemented in the NVIDIA NeMo\footnote{\url{https://github.com/NVIDIA/NeMo}} framework \cite{kuchaiev2019nemo}. They build on baseline sequence-to-sequence transformer models \cite{vaswani2017} in the following ways: 1) Checkpoint averaging, 2) Model scaling up to 1B parameters, 3) Data augmentation with large-scale backtranslation \cite{edunov2018understanding} of monolingual Newscrawl data and sequence-level knowledge distillation from a right-to-left factorized model \cite{zhang2019regularizing}, 4) Finetuning models on in-domain news data from WMT test sets made available in previous years, 5) Ensembling models trained on different subsets of the overall data 6) Shallow fusion decoding with transformer language models \cite{gulcehre2015using} 7) Noisy channel re-ranking of beam search candidate hypotheses \cite{yee2019simple}. 

Overall, we find each of these components results in a small improvement in BLEU scores with backtranslation results being mixed depending on the language direction and whether the test data contains translationese inputs. Using a combination of these techniques, we achieve 39.5 sacreBLEU scores on the En $\rightarrow$ De WMT'20 test set, outperforming the best BLEU scores from last year's competition of 38.77.  

Training our En $\leftrightarrow$ Ru biomedical task submission from scratch using a biomedical vocabulary and similar model improvements to those used for our news task submission, we report a sacreBLEU score of 40.3 on En $\rightarrow$ Ru and 43.8 on Ru $\rightarrow$ Enh on the WMT'20 Biomedical Shared Task test dataset. This improves over the best submissions from last year's competition\footnote{We compare against all En $\leftrightarrow$ Ru Biomedical submissions, not just the ones marked as the final submission.} of 39.6 and 43.3 on En $\rightarrow$ Ru and Ru $\rightarrow$ En respectively. 

\section{Datasets}
We participated in the constrained data track at this year's news and biomedical competitions and used all the parallel corpora provided by the WMT Shared Tasks for both En $\leftrightarrow$ De and En $\leftrightarrow$ Ru. We used the provided English, German, and Russian monolingual Newscrawl data for backtranslation and training our autoregressive transformer language models. We filter out monolingual Newscrawl data only based on minimum and maximum length criteria, but perform more aggressive filtering of our parallel data described in Section \ref{sec:data_filtering}.

\subsection{Parallel Corpus Filtering}
\label{sec:data_filtering}
We use a combination of the following data filtering steps for all parallel corpora (including pseudo parallel corpora generated via backtranslation and distillation) except for the Biomedical Shared Task provided data.

\begin{itemize}
    \item \textbf{Language ID Filtering} - We use the fastText \cite{joulin2016fasttext} language ID classifier\footnote{\url{https://fasttext.cc/docs/en/language-identification.html}} to remove training examples that aren't in the appropriate language.
    \item \textbf{Length and Ratio Filtering} - We filter out examples where a sentence in either language is longer than 250 tokens before BPE tokenization and where the length ratio between source and target sentences exceeds 1.3.
    \item \textbf{Bicleaner} - Bitexts that were assigned a Bicleaner \cite{ramirez2020bifixer} score of < 0.6 were removed.
\end{itemize}

On the news shared task, we keep ~60M parallel sentences for En $\leftrightarrow$ De and ~26M sentences for En $\leftrightarrow$ Ru after filtering.

\subsection{Biomedical Task Data}
\label{sec:biomed-data}
Our parallel biomedical domain data included a mix of all the En $\leftrightarrow$ Ru parallel training data given by shared task organizers and biomedically relevant examples selected from the provided En $\leftrightarrow$ Ru news task data. 

We trained two biomedical domain binary classifiers, one for English and one for Russian. The classifiers were composed of two task-specific fully connected layers on top of pre-trained BERT Base \cite{devlin2018bert} or RuBERT Base \cite{kuratov2019adaptation} for English and Russian respectively. The positive examples were sourced from the WMT'20 Biomedical Shared Task train set. The negative examples were randomly sampled from the parallel En $\leftrightarrow$ Ru news data given for the WMT'21 news task. An equal amount of 45K examples were used for both the positive and negative classes.

We ran our English biomedical domain classifiers on the English half of all approximately 26M parallel En $\leftrightarrow$ Ru WMT'21 news training data. We saved all sentences with predicted biomedical domain probabilities over 50\%, collecting around 560k examples. We then ran our Russian classifier on the Russian counterparts to the 560k predicted in domain English sentences. We averaged the classifier scores from the English and Russian domain classifiers and used this average score as our final selection criteria. We set a cut-off threshold of .90 resulting in 208K parallel examples classified from the news domain data. We combined this with the 46k parallel biomedical examples provided for the task, resulting in a total of 256,037 parallel training examples. 

\subsection{Data Pre-processing and Post-processing}
\label{sec:data_processing}
We normalize punctuation\footnote{\url{https://github.com/moses-smt/mosesdecoder/blob/master/scripts/tokenizer/normalize-punctuation.perl}} and tokenize\footnote{\url{https://github.com/moses-smt/mosesdecoder/blob/master/scripts/tokenizer/tokenizer.perl}} examples with the Moses toolkit. For En $\leftrightarrow$ De, we train a shared BPE tokenizer with a vocab of 32k tokens using the YouTokenToMe\footnote{\url{https://github.com/VKCOM/YouTokenToMe}} library. For En $\leftrightarrow$ Ru, we train language-specific BPE tokenizers with a vocab of 16k tokens each. For the En $\leftrightarrow$ Ru Biomedical translation task, we learn a separate BPE tokenizer solely on our Biomedical Task Data described in \ref{sec:biomed-data}. We use BPE-dropout \cite{provilkov2019bpe} of 0.1 for both language pairs and tasks. We post-process En $\rightarrow$ De model generated translations to replace quotes with their German equivalents - „ and “.

\section{System overview}
Our systems build on the Transformer sequence-to-sequence architecture \cite{vaswani2017}. In the subsequent subsections, we discuss model scaling, checkpoint averaging, data augmentation with backtranslation and right-to-left distillation, model finetuning, ensembling, shallow fusion decoding with LMs, and noisy channel re-ranking.

\subsection{Model Configurations}
We experiment with three different model configurations - Large, XLarge, and XXLarge. The Large configuration corresponds to the ``Transformer Large'' variant from \citet{vaswani2017} and the XLarge and XXLarge scale that base configuration along depth and width. The exact specifications are in Table \ref{tab:model_size}. Following \citet{kasai2020deep}, we keep the number decoder layers fixed at 6 and scale only the depth of the encoder to 24 layers for the ``XLarge'' configuration. For stable optimization of deep transformers, we use the ``pre-LN'' transformer block \cite{xiong2020layer}. When scaling to 1 billion parameters (XXLarge), we only increase hidden and feedforward dimensions of the model.

\begin{table}[h]
    \centering
    \begin{tabular}{|c|c|c|c|}
    \hline
    & Large & XLarge & XXLarge \\
    \hline
    Hidden Dim & 1,024 & 1,024 & 1,536 \\
    Feedforward Dim & 4,096 & 4,096 & 6,144 \\
    Attention Heads & 16 & 16 & 24 \\
    Encoder Layers & 6 & 24 & 24 \\
    Decoder Layers & 6 & 6 & 6 \\
    Pre-LN & \xmark & \cmark & \cmark \\
    \hline
    Parameters & 240M & 500M & 1B \\
    \hline
    \end{tabular}
    \caption{Model Configurations}
    \label{tab:model_size}
\end{table}
\subsection{Checkpoint Averaging}
Over the course of training, we save the top-k checkpoints that obtain the best sacreBLEU scores on a validation set. The final model parameters are obtained by averaging the parameters corresponding to these checkpoints.
$$\theta_{avg} = \frac{1}{k} \sum_{i=1}^{k} \theta_{i}$$

$\theta_{avg}$ are the model parameters after checkpoint averaging and $\theta_{1} \ldots \theta_{k}$ are the individual checkpoints being averaged. Empirically, we didn't observe a difference between averaging the last k checkpoints versus the top-k checkpoints. The former is however more common and implemented in libraries such as fairseq \cite{ott2019fairseq}.

\subsection{Data Augmentation with Backtranslation \& Right-to-left model distillation}
We follow \citet{edunov2018understanding} in backtranslating monolingual Newscrawl data with noise introduced via topk sampling (k=500). For En $\leftrightarrow$ De, we backtranslate \textasciitilde 250M sentences and filter translations based on the process described in Section \ref{sec:data_filtering}. We observed fairly significant drops in BLEU score when using backtranslated data for En $\leftrightarrow$ Ru and did not apply any data augmentation for this language pair. We use the XLarge model configuration trained only on the News Task provided parallel corpus to generate translations.

We also train an XLarge model for En $\rightarrow$ De and De $\rightarrow$ En on the News Task provided parallel data where the output sequence is factorized from right-to-left. Translations of the training dataset with topk sampling (k=500) using these models are generated and added to the overall training set.

When adding only backtranslated text or data generated from right-to-left factorized models, we use a 2:1 ratio of parallel to pseudo-parallel (model generated) data. When training with a combination of both, we use a 6:3:1 ratio of parallel, right-to-left generated, and backtranslated data. We skew data sampling in this way since training on right-to-left generated data showed better performance on recent WMT test sets as opposed to backtranslation which did better on old test sets that contained translationese inputs (see Tables \ref{tab:en_de_ablation} and \ref{tab:de_en_ablation}).

\subsection{Mixed Domain Training}
For the biomedical task submission, we experiment with different mixed domain training approaches \cite{zhang-etal-2019-curriculum}. We train on the concatenated combination of news task and biomedical task data- up-sampling the proportion of biomedical data to make up 30\% or 50\% of the data-parallel examples seen during training. We also train models on concatenated data with no up-sampling and with purely news task data. The base models trained on exclusively news task data still use the biomedical vocabulary tokenizer. 
 
\subsection{Model Finetuning}
For our news task submission, we finetuned models on an in-domain parallel corpus consisting of WMT provided test datasets from past years (WMT'08 - WMT'19 for En $\leftrightarrow$ De comprising \textasciitilde 32k examples) for both En $\leftrightarrow$ De and En $\leftrightarrow$ Ru. 

We finetuned our biomedical task base models on the 250k parallel sentences obtained via the process described in Section \ref{sec:biomed-data}. Models are finetuned for 1-2 epochs using a fixed tuned learning rate and the top-k checkpoints on a validation dataset (newstest2020 for the News Shared Task) are averaged.

\subsection{Ensembling}
Given $k$ different models for a particular language direction trained with the same tokenizer, we ensemble them at inference by averaging their probability distributions over the next token.

$$P(y_t|y_{<t},x;\theta_{1} \ldots \theta_{k}) = \frac{1}{k} \sum_{i=1}^k P(y_t|y_{<t},x;\theta_{i})$$

Where $P(y_t|y_{<t},x)$ is the probability distribution over the target token $y_t$ given all previously generated target tokens $y_{<t}$ and the input sequence $x$. $\theta_{1} \ldots \theta_{k}$ are the $k$ different models being ensembled.

Beam search scores are computed using these averaged probabilities at each time step. In practice, we ensemble models trained on different subsets of the available data.

For En $\leftrightarrow$ De, we ensemble a total of 6 models trained on different subsets of the data. Example: News Task provided bitext only, the addition of backtranslated and/or data from right-to-left factorized models and finetuned models.

For En $\leftrightarrow$ Ru, we ensemble a total of 3 identical XLarge models trained with different random seeds on the main parallel corpus.

For the En $\leftrightarrow$ Ru biomedical task, we ensemble 4 finetuned models whose base configurations were trained with different mixed domain sampling ratios. Specifically, each translation direction includes an ensemble of models initially trained on mixed domain data with 50\% up-sampling of biomedical data, concatenated biomedical and news data with no up sampling, exclusively news data, and exclusively news data with right-to-left distillation. 

\subsection{Shallow Fusion Decoding with Language Models}
Aside from backtranslation, another way to leverage large amounts of monolingual data is via training language models. We train language-specific 16-layer transformer language models at the sentence level, which is architecturally similar to \citet{radford2019language}. They are trained on Newscrawl and use the same tokenizers as our NMT systems. 

When generating translations, we decode jointly with our NMT system $\theta_{s \rightarrow t}$ and a target side language moel $\theta_{t}$ \cite{gulcehre2015using}. The score of a partially decoded sequence on the beam $\mathcal{S}(y_{1 \ldots n})$ of length $n$ is given by the following recurrence

\begin{equation*}
\begin{split}
    \mathcal{S}(y_{1\ldots n}|x;\theta_{s \rightarrow t},\theta_{t}) = \mathcal{S}(y_{1\ldots n - 1}|x;\theta_{s \rightarrow t},\theta_{t}) \\
    + \log P(y_{n}|y_{<n},x;\theta_{s \rightarrow t}) + \lambda_{sf} \log P(y_{n}|y_{<n};\theta_{t})
\end{split}
\end{equation*}

where the empty sequence has a score of 0. We tuned the LM importance coefficient $\lambda_{sf}$ on a validation dataset and found a value between 0.05 - 0.1 to work well in practice.

\subsection{Noisy Channel Re-ranking}
We re-rank the beam search candidates produced by our ensemble model generated with or without shallow fusion using a neural noisy channel model \cite{yee2019simple}. The noisy channel model computes the score of any translation $\mathcal{S}(y_i|x)$ on the beam based on a forward (source-to-target) model, a reverse (target-to-source), and a target language model. The best translation after re-ranking is given by

\begin{equation*}
\begin{split}
     \argmax_{i} \mathcal{S}(y_i|x) = \log P(y_i|x;\theta_{s \rightarrow t}^{ens}) \\ + \lambda_{ncr} \big( \log P(x|y_i;\theta_{t \rightarrow s}) + \log P(y_i;\theta_{t}) \big)
\end{split}
\end{equation*}

Forward log probabilities are given by an ensemble of source-to-target models $\theta_{s \rightarrow t}^{ens}$. We experimented with using an ensemble of target-to-source translation models to compute $\log P(x|y_i)$ but didn't observe any empirical benefits and so all reported results use only a single reverse model $\theta_{t \rightarrow s}$ for noisy channel re-ranking. We generate 15 candidates via beam search and tune $\lambda_{ncr}$ on a validation dataset and found a value between 0.5 - 0.7 to work well in practice.

\subsection{Training \& Optimization}
All En $\leftrightarrow$ De models were trained for up to 450k updates using the Adam optimizer \cite{kingma2014adam} with $\beta_1 = 0.9,\beta_2=0.98$ and Inverse Square Root Annealing \cite{vaswani2017} with 30k warm-up steps and a maximum learning rate of 4e-4. En $\leftrightarrow$ Ru models were trained for up to 150k updates with 7k warmup steps. We use label smoothing of 0.1 and a dropout of 0.1 on intermediate activations including attention scores to regularize our models.

The ``Large'' models were trained on NVIDIA DGX-1 machines with 8 32G V100 GPUs. We use a batch size of 16k tokens per GPU for an effective batch size of 128k tokens. The ``XLarge'' models were trained on 64 GPUs split across 4 NVIDIA DGX-2 nodes with 16 32G V100 GPUs each. These models use an effective batch size of 256k tokens. Finally, our ``XXLarge'' models were trained on 256 GPUs across 16 DGX-2 nodes with an effective batch size of 512k tokens.

\begin{table*}[ht!]
    \centering
    \begin{tabular}{|c|p{5.8cm}|c|c|c|c|c|}
    \hline
    & En $\rightarrow$ De News Task Model & WMT'14 & WMT'18 & WMT'19 & WMT'20 & Avg $\Delta$ \\
    \hline
    (1) & Transformer-Large & 29.9 & 46.6 & 41.1 & 31.5 & 0 \\
    \hline
    (2) & (1) + Checkpoint Averaging & 30.7 & 48.3 & 43.5 & 33.5 & 1.4 \\
    \hline
    (3) & (2) + Transformer-XLarge & 32.2 & 48.7 & 43.3 & 34.7 & 2.1 \\
    \hline
    (4) & (3) + Backtranslation & 34.9 & 49.2 & 40.5 & 34.6 & 2.2 \\
    (5) & (3) + R2L Distllation & 32.4 & 49.1 & 43.4 & 37.2$^*$ & 2.9 \\
    (6) & (3) + Backtranslation + R2L Distllation & 34.3 & 50.1 & 42.9 & 37.4$^*$ & 3.6 \\
    \hline
    (7) & (5) + Shallow Fuison Decoding & 32.8 & 49.0 & 43.4 & 37.6$^*$ & 3.1 \\
    \hline
    (8) & (6) + Transformer-XXLarge & 35.5 & 50.0 & 41.8 & 37.5$^*$ & 3.6 \\
    (9) & (6) + Finetuning (WMT'08-19) & - & - & - & 37.6$^*$ & - \\
    \hline
    (10) & (8) + (9) + Ensembling & 34.4 & 50.7 & 44.2 & 38.9$^*$ & 4.4 \\
    \hline
    (11) & (10) + Noisy Channel Re-ranking & \textbf{36.0} & \textbf{51.6} & \textbf{44.3} & \textbf{39.5}$^*$ & 5.2 \\
    \hline
    \end{tabular}
    \caption{Model ablations for En $\rightarrow$ De. All reported scores are obtained from sacreBLEU. WMT'20 scores with a $^*$ apply post-processing to replace punctuations as reported in Section \ref{sec:data_processing}. Avg $\Delta$ computes the improvement over the Transformer-Large baseline averaged over the 4 test sets.}
    \label{tab:en_de_ablation}
\end{table*}

\begin{table*}[ht!]
    \centering
    \begin{tabular}{|c|p{5.9cm}|c|c|c|c|c|}
    \hline
    & De $\rightarrow$ En News Task Model & WMT'14 & WMT'18 & WMT'19 & WMT'20 & Avg $\Delta$ \\
    \hline
    (1) & Transformer-Large & 35.5 & 45.0 & 40.5 & 37.5 & 0 \\
    \hline
    (2) & (1) + Checkpoint Averaging & 36.5 & 46.1 & 41.6 & 38.3 & 0.7 \\
    \hline
    (3) & (2) + Transformer-XLarge & 37.7 & 47.8 & 41.9 & 37.6 & 1.3 \\
    \hline
    (4) & (3) + Backtranslation & 40.3 & 50.4 & 40.5 & 37.7 & 2.3 \\
    (5) & (3) + R2L Distllation & 37.5 & 47.8 & 42.3 & 39.7 & 1.9 \\
    (6) & (3) + Backtranslation + R2L Distllation & 39.3 & 49.6 & 41.8 & 39.4 & 2.7 \\
    \hline
    (7) & (6) + Finetuning (WMT'08-19) & - & - & - & 41.1 & - \\
    \hline
    (8) & (7) + Ensembling & 39.5 & 49.9 & \textbf{43.3} & 41.9 & 3.7 \\
    \hline
    (9) & (8) + Noisy Channel Re-ranking & \textbf{40.1} & \textbf{50.6} & 42.8 & \textbf{42.0} & 4.0 \\
    \hline
    \end{tabular}
    \caption{Model ablations for De $\rightarrow$ En. All reported scores are obtained from sacreBLEU. Avg $\Delta$ computes the improvement over the Transformer-Large baseline averaged over the 4 test sets.}
    \label{tab:de_en_ablation}
\end{table*}

\begin{table*}[ht!]
    \centering
    \begin{tabular}{|c|p{5.9cm}|c|c|c|c|c|}
    \hline
    & En $\rightarrow$ Ru News Task Model & WMT'17 & WMT'18 & WMT'19 & WMT'20 & Avg $\Delta$ \\
    \hline
    (1) & Transformer-Large & 35.4 & 30.8 & 32.0 & 22.3 & 0 \\
    \hline
    (2) & (1) + Transformer-XLarge + Ckpt Avg & 36.8 & 32.2 & 33.2 & 23.2 & 1.2 \\
    \hline
    (3) & (2) + Finetuning (WMT'14-16) & 38.0 & 33.1 & 35.1 & 24.3 & 2.5 \\
    \hline
    (4) & (3) + Ensemble (x3) &  \textbf{38.6} & 33.5 & 35.3 & \textbf{24.8} & 2.9 \\
    \hline
    (5) & (4) + Shallow Fusion & \textbf{38.6} & \textbf{33.7} & \textbf{35.7} & 24.7 & 3.0 \\
    \hline
    (6) & Oracle BLEU with beam size 4 & - & - & 39.9 & - & - \\
    \hline
    \end{tabular}
    \caption{Model ablations for En $\rightarrow$ Ru. All reported scores are obtained from sacreBLEU. Avg $\Delta$ computes the improvement over the Transformer-Large baseline averaged over the 4 test sets.}
    \label{tab:en_ru_ablation}
\end{table*}

\begin{table*}[ht!]
    \centering
    \begin{tabular}{|c|p{5.8cm}|c|c|c|c|c|}
    \hline
    & Ru $\rightarrow$ En News Task Model & WMT'17 & WMT'18 & WMT'19 & WMT'20 & Avg $\Delta$ \\
    \hline
    (1) & Transformer-Large & 37.6 & 33.0 & 37.7 & 36.6 & 0 \\
    \hline
    (2) & (1) + Transformer-XLarge + Ckpt Avg &  38.7 & 34.3 & 38.2 & 37.2 & 0.9 \\
    \hline
    (3) & (2) + Finetuning (WMT'14-16) & 40.7 & 35.4 & 40.5 & 37.1 & 2.2 \\
    \hline
    (4) & (3) + Ensemble (x3) & 40.7 & 35.5 & \textbf{41} & \textbf{37.7} & 2.5 \\
    \hline
    (5) & (4) + Shallow Fusion & \textbf{40.9} & \textbf{35.9} & 40.8 & 37.5 & 2.6 \\
    \hline
    (6) & Oracle BLEU with beam size 4 & - & - & 46.4 & - & - \\
    \hline
    \end{tabular}
    \caption{Model ablations for Ru $\rightarrow$ En. All reported scores are obtained from sacreBLEU. Avg $\Delta$ computes the improvement over the Transformer-Large baseline averaged over the 4 test sets.}
    \label{tab:ru_en_ablation}
\end{table*}

\begin{table*}[ht!]
    \centering
    \begin{tabular}{|c|l|c|c|}
    \hline
    & En $\rightarrow$ Ru Biomedical Task Model & WMT'20 Bio & $\Delta$ \\
    \hline
    (1) & Transformer-Large News Task Model & 32.2 & 0 \\
    \hline
    (2) & Transformer-XLarge News Task Model & 33.8 & 1.6 \\
    \hline
    (3) & Transformer-XLarge + Biomed Vocab w/ News Data & 33.9 & 1.7 \\
    \hline
    (4) & Transformer-XLarge + Biomed Vocab w/ News + R2L Distillation Data & 34.2 & 2.0 \\
    \hline
    (5) & Transformer-XLarge + Biomed Vocab w/ News + 30\% Biomed Data & 36.7 & 4.5 \\
    \hline
    (6) & Transformer-XLarge + Biomed Vocab w/ News + 50\% Biomed Data & 36.8 & 4.6 \\
    \hline
    (7) & Transformer-XLarge + Biomed Vocab w/ News + Biomed Data & 37.4 & 5.2 \\
    \hline
    (8) & (2) + Biomed Data Finetuning & 37.8 & 5.6 \\
    \hline
    (9) & (3) + Biomed Data Finetuning & 38.5 & 6.3 \\
    \hline
    (10) & (4) + Biomed Data Finetuning & 38.2 & 6.0 \\
    \hline
    (11) & (6) + Biomed Data Finetuning & 37.4 & 5.2 \\
    \hline
    (12) & (7) + Biomed Data Finetuning & 38.5 & 6.3 \\
    \hline
    (13) & (9) (10) (11) (12) Ensemble & 39.9 & 7.7 \\
    \hline
    (14) & (13) + Shallow Fusion & 40.0 & 7.8 \\
    \hline
    (15) & (13) + Noisy Channel Re-ranking & \textbf{40.3} & 8.1 \\
    \hline
    \end{tabular}
    \caption{Model iterations for the Biomedical Shared Task En $\rightarrow$ Ru. All reported scores are checkpoint averaged and are obtained from sacreBLEU. $\Delta$ computes the improvement over the Transformer-Large baseline on the WMT'20 Biomedical Shared Task test set. Model 15 is our selected best submission and model 14 is our alternate submission.}
    \label{tab:en_ru_biomed_iterations}
\end{table*}

\begin{table*}[ht!]
    \centering
    \begin{tabular}{|c|l|c|c|}
    \hline
    & Ru $\rightarrow$ En Biomedical Task Model & WMT'20 Bio & $\Delta$ \\
    \hline
    (1) & Transformer-Large News Task Model & 38.7 & 0 \\
    \hline
    (2) & Transformer-XLarge News Task Model & 39.8 & 1.1 \\
    \hline
    (3) & Transformer-XLarge + Biomed Vocab w/ News Data & 39.8 & 1.1 \\
    \hline
    (4) & Transformer-XLarge + Biomed Vocab w/ News + R2L Distillation Data & 39.2 & 0.5 \\
    \hline
    (5) & Transformer-XLarge + Biomed Vocab w/ News + 30\% Biomed Data & 37.6 & -1.1 \\
    \hline
    (6) & Transformer-XLarge + Biomed Vocab w/ News + 50\% Biomed Data & 38.4 & -0.3 \\
    \hline
    (7) & Transformer-XLarge + Biomed Vocab w/ News + Biomed Data & 41.5 & 2.8 \\
    \hline
    (8) & (2) + Biomed Data Finetuning & 42.3 & 3.6 \\
    \hline
    (9) & (3) + Biomed Data Finetuning & 42.6 & 3.9 \\
    \hline
    (10) & (4) + Biomed Data Finetuning & 41.7 & 3.0 \\
    \hline
    (11) & (6) + Biomed Data Finetuning & 39.6 & 0.9 \\
    \hline
    (12) & (7) + Biomed Data Finetuning & 41.8 & 3.1 \\
    \hline
    (13) & (9) (12) Ensemble & 42.8 & 4.1 \\
    \hline
    (14) & (9) (10) (11) (12) Ensemble & \textbf{43.8} & 5.1 \\
    \hline
    (15) & (14) + Shallow Fusion & 43.7 & 5.0 \\
    \hline
    (16) & (14) + Noisy Channel Re-ranking & 42.1 & 3.4 \\
    \hline
    \end{tabular}
    \caption{Model iterations for the Biomedical Shared Task Ru $\rightarrow$ En. All reported scores are checkpoint averaged and are obtained from sacreBLEU. $\Delta$ computes the improvement over the Transformer-Large baseline on the WMT'20 Biomedical Shared Task test set. Model 14 is our selected best submission.}
    \label{tab:ru_en_biomed_iterations}
\end{table*}

\section{News Task Submission}
In this Section, we present results for our News Shared Task submission. Tables \ref{tab:en_de_ablation} and \ref{tab:de_en_ablation} contain ablations for En $\leftrightarrow$ De and while Tables \ref{tab:en_ru_ablation} and \ref{tab:ru_en_ablation} has ablations for En $\leftrightarrow$ Ru. 

Each of the components we describe improves BLEU scores except for backtranslation and scaling our models to 1B params. Both show mixed results on En $\leftrightarrow$ De - scores improve significantly on WMT'14 and WMT'18 test sets when adding backtranslated data (possibly because these test sets contain translationese inputs) but hurts or does not improve performance on WMT'19 and WMT'20 test sets. Our 1B parameter model does significantly better on WMT'14, but worse on WMT'19 and is comparable to the 500M parameter model on WMT'18 and WMT'20. We found optimization with Adam to be unstable and used AdamW with a weight decay of 0.01 instead. Our final En $\rightarrow$ De model achieves a BLEU score of 39.5 on the  WMT'20 test set, which improves over the submission with the best BLEU score from last year's competition of 38.8. We however do not do as well on De $\rightarrow$ En, with a final BLEU score of 42, compared to last year's best submission of 43.8.

Backtranslation significantly hurt performance in initial experiments on En $\leftrightarrow$ Ru so we exclude it from our ensemble. The impact of ensembling, finetuning, and shallow fusion are fairly similar to En $\leftrightarrow$ De. Additionally, we also report an ``Oracle BLEU'' score in Tables \ref{tab:en_ru_ablation} and \ref{tab:ru_en_ablation} where we compute BLEU scores by cheating and picking the translation on our beam that has the highest sentence BLEU score with respect to the reference. This is a useful indicator of how much there is to gain by re-ranking the beam search candidates.

\section{Biomedical translation task submission}
We present our Biomedical Shared Task submission in this section. Building on lessons learned from our news task ablation studies, we opted to use the Transformer-XLarge architecture, and average all of the intermediate model checkpoints which helps reduce model variance as a consequence of finetuning. Tables \ref{tab:en_ru_biomed_iterations} and \ref{tab:ru_en_biomed_iterations} show our results as we iterated on improving our models.  

We trained our BPE tokenizer on biomedical data to mitigate character-level segmentation of words unique to the biomedical domain. We found this had a minimal effect on model performance. This could be because the majority of our parallel biomedical data was selected from news task training data, and thus biomedical words were already adequately accounted for by the news task model's tokenizer.
We found that up-sampling in-domain biomedical data hurt performance compared to concatenating out-of-domain and in-domain data with no up-sampling. For the En $\rightarrow$ Ru direction, including any biomedical domain data during initial model training showed improvements over training on exclusively news task data. Up-sampling in-domain data for the Ru $\rightarrow$ En direction hurt performance compared to our news task model baselines. 

Unsurprisingly, finetuning base models on biomedical domain data improved BLEU scores for all models. In-domain finetuning helped models initially trained on news task data overcome performance gaps between themselves and models that had seen a higher amount of biomedical data during training. Neither shallow fusion nor noisy channel re-ranking improved model performance after ensembling for the Ru $\rightarrow$ En direction. Both techniques individually improved En $\rightarrow$ Ru performance but failed to do so in combination.  

Ensembling our models led to an additional performance boost and allowed us to reach our maximum En $\rightarrow$ Ru BLEU score of 40.3 and Ru $\rightarrow$ En BLEU score of 43.8. These scores show a 0.9 and 0.5 improvement over last year's best score of 39.4 and 43.3 \cite{bawden-hal} respectively.

\section{Conclusion}
We present Neural Machine Translation Systems for the En $\leftrightarrow$ De News Task and En $\leftrightarrow$ Ru News and Biomedical shared tasks implemented in the NeMo framework \cite{kuchaiev2019nemo}. Our systems build on the Transformer sequence-to-sequence model to include backtranslated text and data from right-to-left factorized models, ensembling, finetuning, mining biomedically relevant data using domain classifiers, shallow fusion with LMs, and noisy channel re-ranking. These achieve competitive performance to submissions from previous years.

\section{Author Contributions}
\paragraph{Sandeep Subramanian:} Sandeep implemented and experimented with model scaling, backtranslation, distillation with right-to-left factorized models, model ensembling, and noisy channel re-ranking. He also ran all of the En $\leftrightarrow$ De News Shared Task experiments, the right-to-left factorized models for the En $\leftrightarrow$ Ru Biomedical task, and wrote parts of the paper.
\paragraph{Oleksii Hrinchuk:} Oleksii H implemented the shallow fusion approach and helped with writing backtranslation scripts. He also ran all of the En $\leftrightarrow$ Ru News Shared Task experiments and trained the language models used in the Biomedical experiments.
\paragraph{Virginia Adams:} Virginia implemented and experimented with the warm-start biomegatron encoder, biomedical baselines, classifiers to extract biomedically relevant monolingual and parallel corpora, mixed domain, and finetuning of News models. She ran all of the En $\leftrightarrow$ Ru Biomedical experiments.
\paragraph{Oleksii Kuchaiev:} Oleksii K advised and managed the project.

\section{Acknowledgements}
The authors would like to thank Mike Chrzanowski, Ryan Prenger, Eric Harper, Micha Livne, Abhinav Khattar, Anton Peganov, Mohammad Shoeybi, Somshubra Majumdar and Fei Jia for many useful discussions over the course of this project.

\bibliography{custom}
\bibliographystyle{acl_natbib}

\appendix



\end{document}